# Fast Training of Deep Networks with One-Class CNNs

Abdul Mueed Hafiz[1*], Ghulam Mohiuddin Bhat[2]

[1, 2] Department of ECE., Institute of
Technology, University of Kashmir
Srinagar, J&K, India, 190006.

[1*]`mueedhafiz@uok.edu.in`

ORC-ID[1]: 0000-0002-2266-3708
ORC-ID[2]: 0000-0001-9106-4699

**Abstract.** One-class CNNs have shown promise in novelty detection. However, very less work has been done on extending them to multiclass classification. The proposed approach is a viable effort in this direction. It uses one-class CNNs i.e., it trains one CNN per class, for multiclass classification. An ensemble of such one-class CNNs is used for multiclass classification. The benefits of the approach are generally better recognition accuracy while taking almost even half or two-thirds of the training time of a conventional multi-class deep network. The proposed approach has been applied successfully to face recognition and object recognition tasks. For face recognition, a 1000 frame RGB video, featuring many faces together, has been used for benchmarking of the proposed approach. Its database is available on request via e-mail. For object recognition, the Caltech-101 Image Database and *17*Flowers Dataset have also been used. The experimental results support the claims made.

**Keywords:** One-Class CNNs; Fast Training; Rapid Convergence; Caltech-101; Deep Learning;

## 1  Introduction

One-class Convolutional Neural Networks (CNNs) are being using for anomaly detection and bi-partitioned space search [1-5]. However, they are not usually applied to multiclass classification. Class-specific Convolutional Neural Networks (CNNs) have been used in action detection and classification in videos. The authors of [6] after drawing inspiration from [7-9] use two Class-specific Networks, the first for RGB images and the second for optical flow images respectively. However, the approach uses two CNNs those too for object detection and not classification by also using dynamic programming. The approach which comes close to the proposed approach is that of [6], which uses cue-based class-specific CNNs  for visual tracking.

However, they use very shallow untrained CNNs in a complex fashion with a small feature vector for use in subsequent stages. Again, in [10] the authors use very shallow CNNs and the length of the feature extracted in the later stages is small which may have affected the volume of contextual information available in the feature. They also initialize their network selectively. The authors also state that the training time is generally higher than those of contemporary techniques.

Deep learning [11-19] frameworks are efficient at extracting discriminative features with increasing amount of context concurrent with increasing number of layers [20]. This has played a role in the use of feature extraction by deep neural networks [21] and subsequent use of these features with various classifiers like Support Vector Machine (SVM) [22], K Nearest Neighbor (K-NN) [23-26], etc. Using a deep neural network (e.g. AlexNet [27]), the features from the fully connected layer(s) are extracted and these are fed to a statistical classifier like SVM or KNN. In R-CNN [28], the authors use class-specific linear SVM classifiers using fixed-length features extracted with the CNN, replacing the softmax classifier learned by fine tuning. However, the positive and negative examples defined for training the SVM classifiers are different from those for fine-tuning the CNN. In [26], the authors use a KNN classifier on top of the last fully connected layer of a deep neural network. The authors do not use the features of the intermediate layers of the deep network. In [24], the authors combine the features of different layers of a deep neural network, and after dimensional reduction apply them to a KNN classifier.

In this paper, a novel approach is proposed which consists of using several one-class CNNs (pre-trained AlexNet's) with Nearest Neighbor (NN) Classifiers. The advantages of this approach are generally better classification accuracy as compared to conventional multiclass deep networks, rapid convergence because of reduced training times and overall simplicity as compared to that of other contemporary techniques, while using the prowess of transfer learning as it benefits from the training experience of powerful CNNs like AlexNet [27]. For every one-class network, training is done as per convention. The networks are used for classification with last three layers removed, whereupon the feature map of the last fully connected layer or 'fc Layer,' is fed to a Nearest Neighbor Classifier. Here, as an experimental choice for demonstration of results, the application of the proposed approach has extended to video face recognition as well as object recognition. An RGB Video (640x480 pixels) of 1000 frames has been used for benchmarking purposes. The benchmarking video is available on request by e-mailing the corresponding author. For object recognition experiments, the Caltech-101 Image Database, and the 17Flowers Dataset have been used. In the experiments, the training time for the proposed technique was found to be as less by almost half or one-third of that taken by other conventional approaches.

## 2    Proposed Approach

The deep network used is AlexNet [27]. Transfer learning is used. The proposed approach uses '*C*' AlexNet's, where '*C*' is the number of classes. Thus it can be said that '*C*' one-class AlexNet's have been used. Let each one-class AlexNet be denoted by *transferNet$_i$*, where $i = 1…C$. First, a separate AlexNet is trained on one class of images. Next, a training dataset *R* comprising of *C* subsets is generated. Each subset $R_i \in R$ consists of the emission of the fully-connected layer or 'fc Layer' of the $i^{th}$ AlexNet i.e. *transferNet$_i$* which is fed on training images for class *i*. For classification of a region-of-interest or *ROI*, each trained AlexNet i.e., *transferNet$_i$* gives its 'fc Layer' emission/feature-map when fed on the *ROI*. Let the Test Set S consist of *C* feature maps: $S_i$, where $i = 1…C$. Next, the distance of the nearest neighbor of $S_1$ in *R* is found and the class of the nearest neighbor is noted. This procedure is repeated for $i=2…C$. This gives a C-row, 2-column array (C×2 array), called **score_db** whose 1$^{st}$ column entries give the minimum distance of $S_i$ to its nearest neighbor in *R,* and whose 2$^{nd}$ column entries give the corresponding class of the nearest neighbor of $S_i$ in *R* respectively. Let $r_m$ be row-index of minimum in 1$^{st}$ column of **score_db**. Thus, $r_m$ corresponds to the least distance among all neighbors of $S_i$ in *R*. Finally, the corresponding 2$^{nd}$ column entry (class of nearest neighbor) of the row having $r_m$ in **score_db** is noted and this is assigned as class of the ROI.

The 'Cosine Metric' distance was used for nearest neighbor search. For an *mx*-by-*n* data matrix *X*, treated as *mx* (1-by-*n*) row vectors $x_1, x_2, ..., x_{mx}$, and an *my*-by-*n* data matrix *Y*, treated as *my* (1-by-*n*) row vectors $y_1, y_2, ..., y_{my}$, the 'cosine distance' $d_{st}$ between the vector $x_s$ and $y_t$ is given by Equation (1).

$$d_{st} = \left(1 - \frac{x_s y'_t}{\sqrt{(x_s x'_s)(y_t y'_t)}}\right) \tag{1}$$

Let $N_{tr}$ be the number of training images per class. The subset $R_i \in R$, consists of $N_{tr} \times 4096$, fc7 Layer emissions of *transferNet$_i$* fed on $N_{tr}$ training images of one class. Thus, the size of the training set is $K \times 4096$ where $K= N_{tr} \times C$, where *C* is the number of classes. The dimension 4096 is used because of the built-in structure of the 'fc Layer' in AlexNet. It should be noted that in the classification stage, the trained AlexNet's are used only up to the 'fc Layer.' Inside the AlexNet architecture, this 'fc Layer' is designated as 'fc7 Layer.' Also, each one-class AlexNet i.e. *transferNet$_i$* is trained end to end on the training images of its class. Figure 1 shows overview of the proposed approach (for four one-class networks).

## 3    Experimentation

The training of the one-class models was done on a machine with an *Intel® Xeon®* (2 core), 16 GB RAM, and 12GB GPU. AlexNet [27] was used as the deep learning model. The input images (ROIs, training images, and testing images) used for the model were RGB images with dimensions 227×227×3. The deep networks were trained using the Stochastic Gradient Descent with Momentum Algorithm, with *Initial Learn Rate* = 0.01, *$L_2$ Regularization Factor* = 0.0001, *Momentum* = 0.9, *Validation Frequency* = 50, and *Validation Patience* = 5. These parameters gave the best results.

The trained deep networks converged successfully in the epochs mentioned below (Further training did not lead to significant improvements in recognition accuracy).

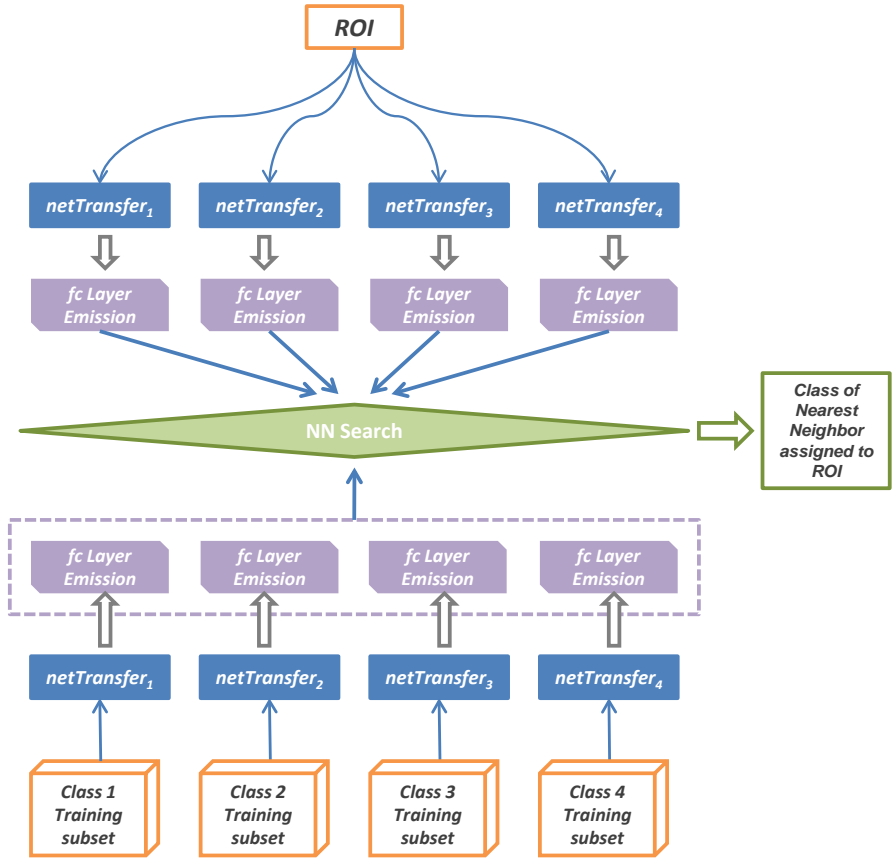

Figure 1. Overview of the proposed approach (for four one-class networks)

### 3.1 *4*Face Database

This database used for benchmarking of the proposed algorithm was a 1000 frame, 640×480 pixels, RGB video and is available on request via e-mail. The video features four persons with varying illuminations and different face orientations due to pan and tilt of head. Figure 2 shows a single frame from the video.

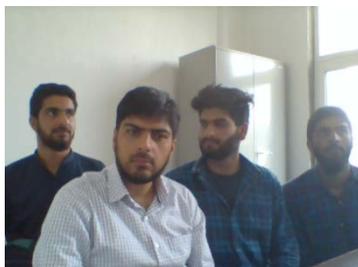

Figure 2. A frame from the benchmarking video

ROIs were extracted per frame, up to a maximum of four. Figure 3 shows the four ROIs extracted from a frame of the video.

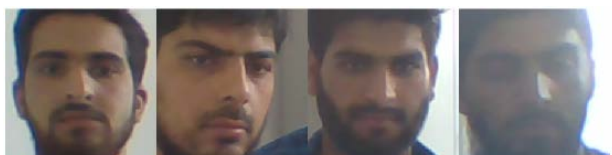

Figure 3. Four ROIs extracted from a frame of the video (Note the varying illumination, which makes the fourth ROI difficult to extract)

The distance metric used in for Nearest Neighbor search was 'cosine distance metric' and the search was conducted for 1 nearest neighbor. This was found to give best recognition accuracy. Since the maximum number of persons is four, hence, four transfer learning trained, one-class AlexNet's were used for the experimentation of the proposed approach. For benchmarking of the performance of the proposed approach, the same is compared with that of a Multiclass AlexNet trained on all four classes of persons using transfer learning, here referred to as *netTransfer4*, and also the Deep Network – KNN Hybrid approach of [24]. The performance of the approaches is shown in Table 1.

**Table 1.** Comparison of various approaches on the *4*Face Database.

| | Person 01 | Person 02 | Person 03 | Person 04 | Test Accuracy (%) | Total Deep Network Training Time (sec) |
|---|---|---|---|---|---|---|
| **Total ROIs in Test Set** | 243 | 261 | 261 | 153 | - | - |
| **ROIs correctly recognized by conventional Deep Learning Approach** | 195 | 261 | 261 | 153 | 94.7 | 391.14 |

| | | | | | |
|---|---|---|---|---|---|
| **(Using a Single four-class AlexNet trained by Transfer Learning)** <br> TrainingSet=140images× 4classes=560 images <br> Test Set=60images× 4classes=240 images <br> Mini-batch size = 16; Epochs=1; Iterations = 35; | | | | | | |
| **ROIs correctly recognized by Proposed Approach (Using 4 one-class networks with KNN)** <br> Training set per Network = 80 images <br> Test Set per Network = 40 images <br> Mini-batch size = 40; Epochs=1;Iterations =2;K = 1; | 216 | 261 | 261 | 153 | 97.1 | **222.01** |
| **ROIs correctly recognized by approach of [24] (Using a Single four-class AlexNet trained by Transfer Learning with KNN)** <br> TrainingSet=140images×4clas ses=560 images <br> Test Set=60images×4classes=240 images <br> Mini-batch size = 16; Epochs=1;Iterations=35;K= 1; | 224 | 261 | 261 | 153 | 97.9 | 391.14 |

All networks are trained on parameters which give sufficient training. As is observable from Table 1, the recognition accuracy of the proposed approach is better than that of a conventional deep learning model (Multi-class AlexNet trained for four classes using transfer learning), while coming next to the recognition accuracy of the approach of [24]. However, the total training time taken (using transfer learning) by all the four one-class deep networks is almost half of that taken by a four-class AlexNet. This is an advantage of the proposed approach over other approaches given in Table 1. As is evident from Table 1, the proposed approach gives comparable recognition accuracy to other conventional techniques in spite of the fact that the one-class networks are trained on a much smaller number of images (Training images per Network = 80, Testing images per Network = 40).

For fine-tuning the performance of the proposed approach, various distance metrics were used in the nearest neighbor search algorithm. The performance of the proposed approach is shown in Table 2.

**Table 2**. Performance of proposed approach for various distance metrics (K=1, Total number of sample frames = 40, Person to recognize = '02')

| Metric | Cosine | Correlation | Spearman |
|---|---|---|---|
| **Number of Frames in which Person 02 was correctly recognized** | **29*** | 22 | 24 |

Using cosine distance metric with K=1 i.e. one nearest neighbor gave best results. The processing time for an ROI using the proposed approach is slightly more than that of the conventional approach using Deep Learning and KNN.

### 3.2 Caltech-101 Image Database

Caltech-101 Image Database [29] was also used for evaluation of the proposed technique. Figure 4 shows some images from the database.

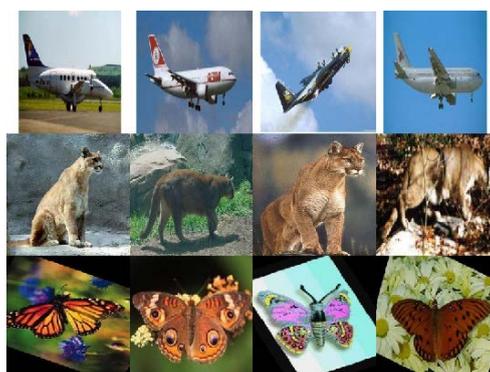

Figure 4. Some images from Caltech-101 Image Database for three categories ('airplanes', 'cougar_body', 'butterfly')

Fifty categories were selected. The number of images selected from each category was 40. Thus equality in number of images per category was used. The images were randomly selected. The results of the experiments are shown in Table 3.

**Table 3.** Comparison of various approaches on Caltech-101 Image Dataset.

| Approach | Test Accuracy (%) | Total Deep Network Training Time (seconds) |
|---|---|---|
| **Conventional Deep Learning Approach (Using a Single fifty-class AlexNet trained by Transfer Learning)**<br>Training set = 15 images × 50 classes = 750 images<br>Test Set = 25 images × 50 classes = 1250 images<br>Mini-batch size = 15; Epochs = 1;Iterations = 50 | 85.2 | 597.52 |
| **ROIs correctly recognized by Proposed Approach Using fifty one-class networks with KNN Classifier)**<br>Training set per Network = 15 images<br>Test Set per Network = 25 images<br>Mini-batch size = 15; Epochs = 1; Iterations = 1; K = 1 | 86.6 | **407.25** |
| **Approach of [24]**<br>**Using a Single fifty-class AlexNet trained by Transfer Learning with KNN)**<br>Training set = 15 images × 50 classes = 750 images<br>Test Set = 25 images × 50 classes = 1250 images<br>Mini-batch size = 15;Epochs = 1; Iterations = 50; K = 1 | **87.3** | 597.52 |

Note that the training time of proposed approach is much lesser as compared to those of other techniques.

### 3.3    *17*Flowers Database

*17*Flowers Image Database [30] was also used fully for evaluation of the proposed technique. Figure 5 shows some images from the database. All images in the available 17 categories were selected. The number of images for each category in the database is 80. For experimentation the images were randomly selected with 70% of images (i.e. fifty-six images) for training and 30% of images (i.e. twenty-four images) for testing per class. The results of the experiments are shown in Table 4.

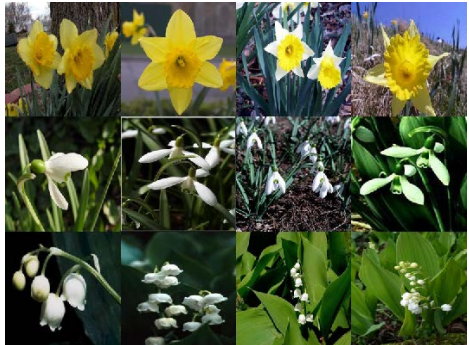

Figure 5. Some images from *17*Flowers Dataset

**Table 4.** Comparison of various approaches on *17*Flowers Dataset.

| Approach | Test Accuracy (%) | Total Deep Network Training Time (seconds) |
|---|---|---|
| **Conventional Deep Learning Approach** **(Using a Single seventeen-class AlexNet trained by Transfer Learning)** Training set = 56 images × 17 classes = 952 images Test Set = 24 images × 17 classes = 408 images Mini-batch size = 56; Epochs = 1; Iterations = 17; | 92.2 | 774.82 |
| **ROIs correctly recognized by Proposed Approach** **(Using seventeen one-class networks with KNN)** Training set per Network = 56 images Test Set per Network = 24 images Mini-batch size=56;Epochs=1;Iterations=1;K = 1; | 93.6 | **543.32** |
| **Approach of [24]** **(Using a Single seventeen-class AlexNet trained by Transfer Learning with KNN)** Training set = 56 images × 17 classes = 952 images Test Set = 24 images × 17 classes = 408 images Mini-batch size=56;Epochs=1;Iterations=17;K = 1; | **94.9** | 774.82 |

Note the much lesser training time of proposed approach as compared to those of other techniques.

## 4    Conclusion and Future Work

In the context of one-class CNNs, very little work has been done on using them for multi-class classification. In this paper, a novel deep learning technique is presented for multi-class classification, which consists of using one-class deep networks combined with KNN classification. The one-class deep networks are trained only on one class, leading to much lesser training times viz. almost half of that of conventional networks and generally better recognition accuracy. The deep neural network used is AlexNet. For experimentation, we have used the tasks of face recognition and object recognition respectively as applications of the proposed approach. A new face recognition video database has been used which is available on request by email. Also, for the object recognition task, the Caltech-101 Image Dataset, and the *17*Flowers Dataset have been used respectively. The performance of the proposed approach is compared experimentally with that of other conventional approaches and advantages of the proposed approach are noted. Future work would involve making the proposed approach more efficient. Also, efforts will be made to apply the proposed approach to other contemporary network architectures, and also to other deep learning associated tasks like Instance Segmentation [31].

## References

1. Oza P, Patel VM Active Authentication using an Autoencoder regularized CNN-based One-Class Classifier. In: 2019 14th IEEE International Conference on Automatic Face & Gesture Recognition (FG 2019), 14-18 May 2019 2019. pp 1-8. doi:10.1109/FG.2019.8756525

2. Perera P, Patel VM (2019) Learning Deep Features for One-Class Classification. IEEE Transactions on Image Processing 28 (11):5450-5463. doi:10.1109/TIP.2019.2917862

3. Sabokrou M, Khalooei M, Fathy M, Adeli E Adversarially learned one-class classifier for novelty detection. In: Proceedings of the IEEE Conference on Computer Vision and Pattern Recognition, 2018. pp 3379-3388

4. Ruff L, Vandermeulen R, Goernitz N, Deecke L, Siddiqui SA, Binder A, Müller E, Kloft M (2018) Deep One-Class Classification. Paper presented at the Proceedings of the 35th International Conference on Machine Learning, Proceedings of Machine Learning Research,

5. Zhang M, Wu J, Lin H, Yuan P, Song Y (2017) The Application of One-Class Classifier Based on CNN in Image Defect Detection. Procedia Computer Science 114:341-348. doi:https://doi.org/10.1016/j.procs.2017.09.040

6. Li H, Li Y, Porikli F DeepTrack: Learning Discriminative Feature Representations by Convolutional Neural Networks for Visual Tracking. In: BMVC, 2014.

7. Gkioxari G, Malik J Finding action tubes. In: CVPR, 2015.

8. Gemert JCv, Jain M, Gati E, Snoek CG Apt: Action localization proposals from dense trajectories. In: BMVC, 2015. p 4

9. Weinzaepfel P, Harchaoui Z, Schmid C Learning to track for spatio-temporal action localization. In: CVPR, 2015.

10. Lu J, Wang G, Deng W, Moulin P, Zhou J Multi-Manifold Deep Metric Learning for Image Set Classification. In: CVPR, 2015.

11. Goodfellow I, Bengio Y, Courville A (2016) Deep learning. MIT press,

12. Schmidhuber J (2015) Deep learning in neural networks: An overview. Neural networks 61:85-117

13. LeCun Y, Bengio Y, Hinton G (2015) Deep learning. nature 521 (7553):436

14. Hafiz AM, Bhat GM A Survey of Deep Learning Techniques for Medical Diagnosis. In, Singapore, 2020. Information and Communication Technology for Sustainable Development. Springer Singapore, pp 161-170

15. Wang Z, Chen J, Hoi SCH (2020) Deep Learning for Image Super-resolution: A Survey. IEEE Transactions on Pattern Analysis and Machine Intelligence:1-1. doi:10.1109/TPAMI.2020.2982166

16. Zhang Z, Cui P, Zhu W (2020) Deep Learning on Graphs: A Survey. IEEE Transactions on Knowledge and Data Engineering:1-1. doi:10.1109/TKDE.2020.2981333

17. Liu L, Ouyang W, Wang X, Fieguth P, Chen J, Liu X, Pietikäinen M (2020) Deep

Learning for Generic Object Detection: A Survey. International Journal of Computer Vision 128 (2):261-318. doi:10.1007/s11263-019-01247-4

18. Dargan S, Kumar M, Ayyagari MR, Kumar G (2019) A Survey of Deep Learning and Its Applications: A New Paradigm to Machine Learning. Archives of Computational Methods in Engineering. doi:10.1007/s11831-019-09344-w

19. Hafiz AM, Bhat GM (2020) Multiclass Classification with an Ensemble of Binary Classification Deep Networks. arXiv preprint arXiv:200701192

20. Zeiler MD, Fergus R Visualizing and understanding convolutional networks. In: European conference on computer vision, 2014. Springer, pp 818-833

21. Yosinski J, Clune J, Bengio Y, Lipson H How transferable are features in deep neural networks? In: Advances in neural information processing systems, 2014. pp 3320-3328

22. Tang Y (2013) Deep learning using linear support vector machines. arXiv preprint arXiv:13060239

23. Sitawarin C, Wagner D (2019) On the Robustness of Deep K-Nearest Neighbors. arXiv preprint arXiv:190308333

24. Papernot N, McDaniel P (2018) Deep k-nearest neighbors: Towards confident, interpretable and robust deep learning. arXiv preprint arXiv:180304765

25. Le L, Xie Y, Raghavan VV Deep Similarity-Enhanced K Nearest Neighbors. In: 2018 IEEE International Conference on Big Data (Big Data), 10-13 Dec. 2018 2018. pp 2643-2650. doi:10.1109/BigData.2018.8621894

26. Ren W, Yu Y, Zhang J, Huang K Learning Convolutional Nonlinear Features for K Nearest Neighbor Image Classification. In: 2014 22nd International Conference on Pattern Recognition, 24-28 Aug. 2014 2014. pp 4358-4363. doi:10.1109/ICPR.2014.746

27. Krizhevsky A, Sutskever I, Hinton GE Imagenet classification with deep convolutional neural networks. In: Advances in neural information processing systems, 2012. pp 1097-1105

28. Girshick R Fast R-CNN. In: 2015 IEEE International Conference on Computer Vision (ICCV), 7-13 Dec. 2015 2015. pp 1440-1448. doi:10.1109/ICCV.2015.169

29. Fei-Fei L, Fergus R, Perona P Learning generative visual models from few training examples: An incremental bayesian approach tested on 101 object categories. In: 2004 conference on computer vision and pattern recognition workshop, 2004. IEEE, pp 178-178

30. Nilsback M-E, Zisserman A A Visual Vocabulary for Flower Classification. In: CVPR, 2006.

31. Hafiz AM, Bhat GM (2020) A survey on instance segmentation: state of the art. International Journal of Multimedia Information Retrieval. doi:10.1007/s13735-020-00195-x